\documentclass[sigconf, nonacm]{acmart}
\acmConference[GRAIL @ CIKM '26]{GRAIL 2026: Workshop on Generative, Retrieval-augmented, and Agentic Intelligence for Personalization}{November 8, 2026}{Rome, Italy}

\usepackage{listings}
\usepackage{xcolor}
\usepackage{makecell}
\usepackage{float}
\usepackage{tcolorbox}
\usepackage{enumitem}
\usepackage{bbm}
\usepackage{balance}
\AtBeginDocument{%
  }

\usepackage[linesnumbered,ruled,vlined]{algorithm2e}

\begin{document}
\settopmatter{printacmref=true}
\acmDOI{}
\acmISBN{}
\title[OptiCover]{OptiCover: A Scalable and Multi-Surface Framework for Cold-Start Visual Optimization under Semantic Constraints}



\author{Guillaume Guy}
\affiliation{\institution{Airbnb}\city{San Francisco}\country{USA}}
\email{guillaume.guy@airbnb.com}

\author{Mihajlo Grbovic}
\affiliation{\institution{Airbnb}\city{San Francisco}\country{USA}}
\email{mihajlo.grbovic@airbnb.com}

\author{Joe Mills}
\affiliation{\institution{Airbnb}\city{San Francisco}\country{USA}}
\email{joe.mills@airbnb.com}

\author{Selen Uguroglu}
\affiliation{\institution{Airbnb}\city{San Francisco}\country{USA}}
\email{selen.uguroglu@airbnb.com}

\author{Liwei He}
\affiliation{\institution{Airbnb}\city{San Francisco}\country{USA}}
\email{liwei.he@airbnb.com}

\author{Pratiksha Kadam}
\affiliation{\institution{Airbnb}\city{San Francisco}\country{USA}}
\email{pratiksha.kadam@airbnb.com}

\author{Nalini Kartha}
\affiliation{\institution{Airbnb}\city{San Francisco}\country{USA}}
\email{nalini.kartha@airbnb.com}

\author{Qian Aria Li}
\affiliation{\institution{Airbnb}\city{San Francisco}\country{USA}}
\email{aria.li.com}


\begin{teaserfigure}
  \centering
  \includegraphics[width=\textwidth, height=4.9cm, keepaspectratio]{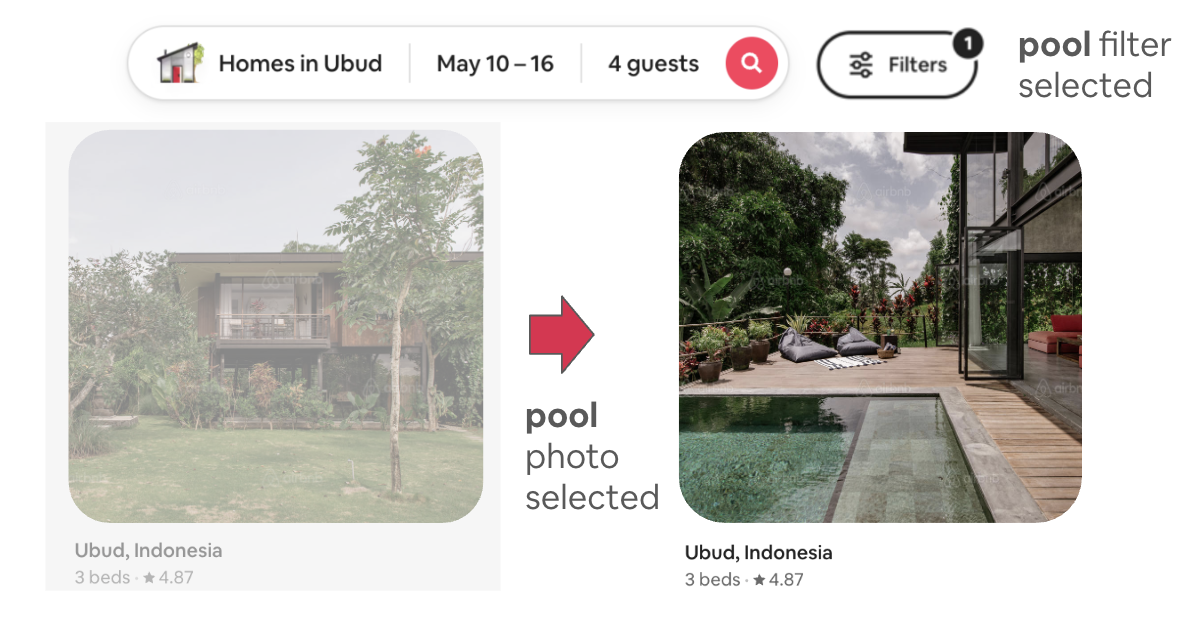}
  \caption{Airbnb Search with Pool as filter before and after applying OptiCover}
  \label{fig:teaser}
  \Description{Airbnb Search with Pool as filter}
\end{teaserfigure}

\begin{abstract}
Selecting an effective cover photo is a critical yet understudied problem in visual merchandising for e-commerce marketplaces. On Airbnb, hosts upload many photos depicting different rooms, amenities, and points of interest, but must manually choose a single static cover photo that is reused across heterogeneous product surfaces. This creates substantial challenges: over 90\% of listing photos have no engagement labels, leading to a severe cold-start regime; guest preferences vary across surfaces and search intents, and visually captivating photos may be semantically misaligned with the underlying listing. At Airbnb scale, this corresponds to hundreds of millions of photos and hundreds of thousands of new uploads per day.

We propose \emph{OptiCover}, a scalable framework for cold-start visual optimization under semantic constraints and multi-surface engagement signals. The framework combines (i) a photo candidate generator that identifies candidate photos through various proxies, (ii) lightweight vision heads applied to frozen CLIP embeddings for large-scale semantic scoring, (iii) a cross-modal grounding validator based on a vision-language model (VLM) that enforces semantic representativeness, and (iv) an Exploration Logging Module (ELM) that orchestrates the collection of randomized engagement data across heterogeneous product surfaces and contextual search intents.

We evaluate OptiCover through extensive offline studies and multiple online experiments across HomePage, and Search surfaces. OptiCover is now deployed in production and performs tens of millions of photo swaps daily. Online results show a 1.5\% engagement lift on the HomePage and a 0.35\% increase in bookers on Search, generating millions of dollars in incremental revenue. Our findings suggest that action-space shaping and semantic constraints are primary drivers of performance in large-scale cold-start visual optimization.
\end{abstract}



\renewcommand{\shortauthors}{Guy et al.}
\maketitle
\makeatletter
\fancypagestyle{standardpagestyle}{%
  \fancyhf{}%
  \renewcommand{\headrulewidth}{\z@}%
  \renewcommand{\footrulewidth}{\z@}%
  \fancyhead[LE,RO]{\@headfootfont\acmConference@shortname,
    \acmConference@date, \acmConference@venue}%
  \fancyhead[LO]{\@headfootfont\shorttitle}%
  \fancyhead[RE]{\@headfootfont\@shortauthors}%
}
\pagestyle{standardpagestyle}
\makeatother

\section{Introduction}
\label{section:introduction}

Cover photos play a central role in online marketplaces, often serving as the first visual signal users encounter. Most platforms delegate the choice of this photo to sellers, despite the fact that items typically have many images that highlight different attributes, spaces, and amenities. On Airbnb, for example, hosts upload dozens of photos but must manually select a single static cover photo that is shown across all user experiences. This choice is inherently subjective and insensitive to user intent: the photo that performs well on the HomePage may not be the best choice for guests searching for a specific amenity (e.g., pools) or visiting during different seasons.

Selecting an ``optimal'' cover photo is therefore a deeply context-dependent problem. Yet, expecting hosts to choose effective photos for all contexts is infeasible. As a result, fixed cover photos frequently fail to reflect what users are searching for, leaving significant engagement and booking value unrealized.

Optimizing cover photos dynamically and at scale introduces several technical challenges. First, marketplaces face a significant cold-start regime: over 90\% of Airbnb listing photos have never been displayed as a cover photo and thus lack direct engagement labels. Second, guest preferences vary substantially across heterogeneous product surfaces (HomePage, Search results), each with distinct visibility, traffic patterns, and user intent. Third, visual optimization must remain semantically faithful to the underlying listing; while some photos are highly engaging, they may depict nearby points of interest or lifestyle scenes that misrepresent the space and risk degrading user on-trip experience. In such environments, the default action space is severely constrained to the current listing covers. Finally, the system must adapt to non-stationary engagement patterns driven by seasonality and demand shifts.

Addressing these challenges requires a framework that can (i) construct a high-quality, both representative and engaging, action space despite severe label sparsity, (ii) enforce semantic constraints to prevent misrepresentation, and (iii) leverage multi-surface engagement signals to generalize across contexts. In this work, we present \emph{OptiCover}, a scalable framework for cold-start visual optimization under semantic constraints and multi-surface engagement signals. OptiCover combines lightweight vision heads trained on frozen CLIP embeddings for large-scale candidate generation, a cross-modal grounding module that validates semantic representativeness, and an Exploration Logging Module that explores photos across heterogeneous surfaces and search intents.

We evaluate OptiCover through extensive offline analyses and multiple online experiments across HomePage and Search surfaces. Beyond improving business metrics, our results highlight broader methodological insights: action-space shaping is a primary driver of performance in a semantically constrained environment, semantic grounding is essential for safe visual optimization, and cross-surface transfer enabling new surface bootstrapping. Together, these findings illustrate a practical and generalizable approach for large-scale visual merchandising in modern marketplaces.

\subsection{Background}
\label{sec:background}

In this section, we outline important prerequisites for understanding the unique challenges associated with our problem at hand, which motivated our approach and architecture choices.

\subsubsection{Data Collection and Seed Training Set}

Modeling photo-driven engagement has historically been challenging due to the lack of large-scale, unbiased training data. In practice, two main approaches have been used, both with significant limitations.

The first approach relies on supervised learning—typically logistic regression—to predict engagement signals such as click-through rate (CTR), using observable confounders as covariates. However, this strategy is highly sensitive to unobserved confounding and selection bias, leading to unstable estimates and limited generalization when applied to ranking or optimization tasks.

The second approach exploits natural experiments, for example, when hosts manually change the cover photo of a listing. Although appealing in principle, this method suffers from several practical issues in practice. Photo changes are often accompanied by simultaneous updates to other listing attributes (e.g., pricing, availability, or description), introducing co-action bias. In addition, engagement signals are strongly affected by temporal factors such as seasonality and demand fluctuations, further complicating causal attribution.

\subsubsection{Cold Start and Shifting Engagement Patterns}

Engagement-based models also suffer from severe cold-start issues and non-stationary engagement patterns. At scale, this problem is particularly acute: every day, hosts upload hundreds of thousands of new photos to their listings. The vast majority of these images have never been used as a cover photo; in practice, we observe over 90\% of listing photos lack any direct engagement observations, making supervised learning infeasible.

In addition, engagement signals on Airbnb are highly seasonal and context-dependent. The relevance of a photo can change substantially over time as guest travel patterns shift with seasonality. For example, mountain destinations such as Lake Tahoe experience significantly different engagement dynamics across seasons, where a winter-focused photo may underperform during summer travel periods, and vice versa. As a result, historical engagement labels can quickly become stale and misaligned with current user preferences.

\subsubsection{Representativeness}

Listing photos convey more than the physical attributes of the property itself. Hosts frequently include images of the surrounding neighborhood, nearby amenities, and local points of interest (POIs) as part of their visual storytelling. While such images can be informative, they introduce an additional constraint when optimizing for engagement: the selected cover photo must remain representative of the actual listing.

Optimizing solely for engagement without enforcing representativeness can result in misleading visual merchandising, where highly engaging but off-premise or contextually ambiguous images are promoted. This, in turn, may negatively impact guest expectations and downstream trip quality, even if short-term metrics such as click-through rates improve.

\subsection{Our Contribution}
\label{sec:our-contribution}

We present \emph{OptiCover}, a scalable framework for cold-start visual optimization under semantic and contextual constraints in a large travel marketplace. To our knowledge, this is one of the first end-to-end systems that optimizes listing cover photos across heterogeneous product surfaces while addressing the combined challenges of sparse supervision, representativeness, and shifting user intent. Our key contributions are:

\begin{itemize}[noitemsep, topsep=0pt]

    \item \textbf{A scalable framework for cold-start visual optimization.}
    We design an end-to-end system that can rank and select cover photos despite the fact that most listing images lack direct engagement labels, enabling robust performance under significant cold-start conditions.

    \item \textbf{A context-aware ranking approach that adapts to user intent and product surface.}
    We introduce a surface- and context-sensitive ranking pipeline that conditions photo eligibility and engagement scoring on search filters, page theme, and surface-level objectives.

    \item \textbf{A lightweight, multi-task vision-head architecture for large-scale semantic scoring.}
    We develop a modular set of vision heads on top of frozen CLIP embeddings to extract engagement predictions, quality signals, and semantic attributes at billion-scale with low latency.

    \item \textbf{A cross-modal grounding module enforcing semantic representativeness.}
    We incorporate a VLM-based grounding validator that filters out non-representative or potentially misleading photos, ensuring optimization does not degrade user trust or trip quality.

    \item \textbf{A practical methodology for deploying visual optimization across heterogeneous product surfaces.}
    We demonstrate that action-space expansion, semantic constraints, surface transfer and calibrated offline predictions are sufficient for safe and high-impact deployment across HomePage, and Search surfaces, yielding significant engagement and booking gains in production.

\end{itemize}

\section{Related Work}
\label{sec:related-work}

Our work intersects three areas: visual merchandising in marketplaces, vision-based engagement modeling, and large-scale online experimentation.

\subsection{Creative Selection and Visual Merchandising}

Prior systems for optimizing representative images—such as Netflix artwork personalization~\cite{amat18}, Tripadvisor cover photo ranking~\cite{Myeongjin25}, and Hotels.com hero-image testing~\cite{Fedor20}—demonstrate that visual choices strongly affect engagement. However, these approaches typically operate over small, well-defined, candidate sets with dense supervision and do not address representativeness constraints or significant cold start. Related work on e-commerce image ranking~\cite{gomez22} and creative generation~\cite{czapp24} emphasizes personalization and aesthetic quality but is less applicable to travel, where photos must remain faithful to the underlying property. In contrast, OptiCover tackles a substantially larger and noisier action space while enforcing semantic fidelity across heterogeneous product surfaces.

\subsection{Vision-Based Engagement Modeling}

Large-scale vision–language models such as CLIP~\cite{radford21} and SigLIP~\cite{tschannen25} enable effective specialization learning for multimodal tasks. Lightweight heads trained on frozen embeddings provide efficient deployment across many classification tasks, and product quantization techniques~\cite{Ge13} support large-scale storage and retrieval~\cite{guy25}. OptiCover builds on these ideas by combining shared frozen embeddings with multi-task vision heads to extract engagement, quality, and semantic signals at scale, and by integrating cross-modal grounding to ensure representativeness.

\subsection{Large-Scale Online Experimentation and Policy Serving}

Hybrid offline–online systems that warm-start online policies with model predictions~\cite{wang21} are effective when candidate sets are compact and rewards are stable. Bandit approaches such as Thompson Sampling have been applied to creative selection~\cite{amat18, Fedor20}, but they assume dense feedback and low-variance reward distributions. In large travel marketplaces, action spaces are large but unvetted, supervision is sparse, and engagement is non-stationary across surfaces and seasons. OptiCover therefore emphasizes offline priors, action-space shaping, and semantic constraints as the primary optimization drivers, using online experimentation chiefly for randomized data collection and policy validation.

\section{Preliminaries}
\label{sec:preliminary}

\begin{figure*}[t]
  \centering
  \includegraphics[width=0.65\linewidth]{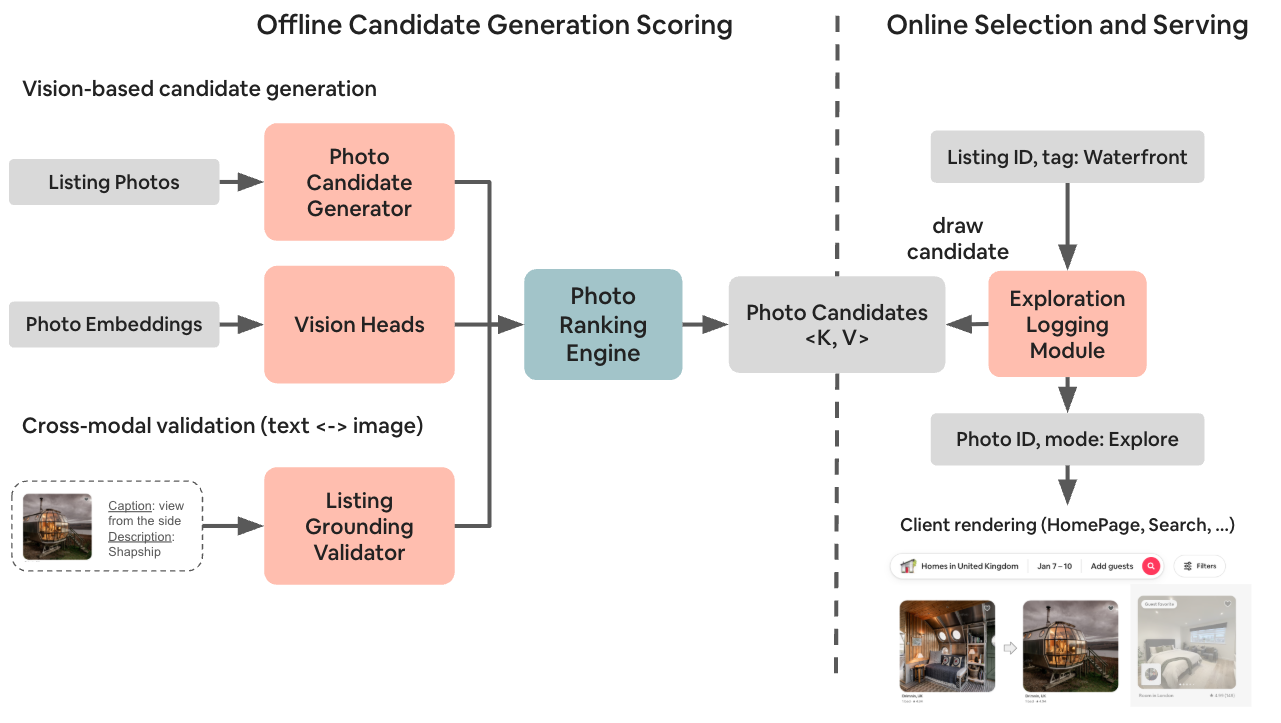}
  \caption{OptiCover architecture
}
  \Description{OptiCover architecture}
  \label{figure:opticover-architecture}
\end{figure*}

\subsection{Problem Setup}

We formalize the cover photo selection task as a contextual optimization problem defined over listing-specific candidate sets.

\subsubsection{Context}
A context $\mathbf{c}$ captures external conditions under which a cover photo is evaluated. In our setting, it is defined as a topical intent $t$ derived from supply-side filters (e.g., \texttt{pool}, \texttt{hot tub}, \texttt{lakefront}). This context determines which photo candidates are eligible and how engagement preferences shift.

\subsubsection{Product Surface}
A product surface $\mathbf{s}$ refers to a specific screen in the guest journey (e.g., HomePage, Search results). Each surface induces distinct traffic patterns, visibility, and optimization objectives.

\subsubsection{Optimization Objective}

Given a business metric of interest—either engagement $E$ or bookers $B$—our goal is to select the cover photo that maximizes expected performance under context $\mathbf{c}$. Formally, for a listing with candidate set $\mathcal{X}$, we solve:
\begin{equation}
\text{find } \mathbf{x}^* =
\arg\max_{\mathbf{x} \in \mathcal{X}} f(\mathbf{x}, \mathbf{s}, \mathbf{c}),
\end{equation}
where $f(\mathbf{x}, \mathbf{s}, \mathbf{c})$ denotes the expected metric (e.g., predicted CTR or booking probability) associated with candidate $\mathbf{x}$ on surface $\mathbf{s}$ with context $\mathbf{c}$. 

\section{Approach}
\label{sec:approach}
\subsection{Objective}

OptiCover aims to discover the most engaging photo for a given search context (or absence of it, in case of the HomePage). Starting from a list of photo cover candidates that passed the quality and representativeness bar, for each listing our goal was to test them on real guest traffic and use the actual guest engagement signals to pinpoint the best one for a given search context. This required building several components of our system that are outlined below.

\subsection{Architecture}
\label{sec:architecture}

OptiCover is a modular architecture designed to generate, score, and validate cover photo candidates in static and context-aware settings. The system consists of the following components:

\begin{itemize}[noitemsep, topsep=0pt]

    \item \textbf{Vision Heads}: A lightweight ML framework that applies multiple task-specific classifiers on top of frozen vision embeddings to estimate engagement and semantic signals at scale.

    \item \textbf{Candidate Generation Engine}: Creates a pool of photo candidates for cover photos for each listing.

    \item \textbf{Listing Grounding Validator}: A VLM-based cross-modal validation module that assesses whether a candidate photo is representative of the underlying listing by comparing visual content against textual and structured listing metadata.

    \item \textbf{Candidate Ranking Engine}: Aggregates vision heads scores and grounding signals to produce a ranked set of photo candidates, along with surface- and context-specific eligibility constraints suitable for online deployment.

    \item \textbf{Exploration Logging Module (ELM)}: Orchestrates explore - exploit across product surfaces, enabling continuous learning and adaptation to shifting engagement patterns.
\end{itemize}

\subsection{Vision Heads}
\label{sec:vision-heads}

This work focused on two general classes of image models: 

\begin{itemize}
    \item \emph{A set of classifiers} to enforce Airbnb editorial policies on visual appearances such as the absence of watermarks, human faces, and certain room types.
    \item \emph{An engagement model} to estimate image click through rate (CTR) as our image selection exploit model.
\end{itemize}
 
 All vision heads are shallow MLPs applied to a frozen ViT CLIP backbone, consisting of a linear layer, a nonlinearity, and a task-specific output layer. Specifically, SigLIP-2 \cite{tschannen25} was selected for the frozen vision backbone. A common frozen ViT backbone was used instead of a fully finetuned ViT for each classification task due to performance constraints of nearly 1 billion listing photos. Due to its wide applicability, a single embedding from the shared ViT can serve all tasks. Furthermore, SigLIP-2 already produces a very strong image representation and Airbnb images are generally within the domain of the large-scale training data used in SigLIP-2. An additional performance optimization was needed because SigLIP-2 produces embeddings of dimension 1152, which can incur substantial latency and storage costs for Airbnb given the large number of images on the platform. All embeddings are quantized to 288 int8 codes with recipe $OPQ288\_1152,PQ288x8$ following the procedure described in \cite{guy25} to reduce storage size, improve iteration speed. During training and inference, we decode the quantized embeddings on the fly back to the full 1152-dimensional vectors by integrating the decoder into the PyTorch forward pass.

\begin{figure}
    \centering
    \includegraphics[width=0.9\linewidth]{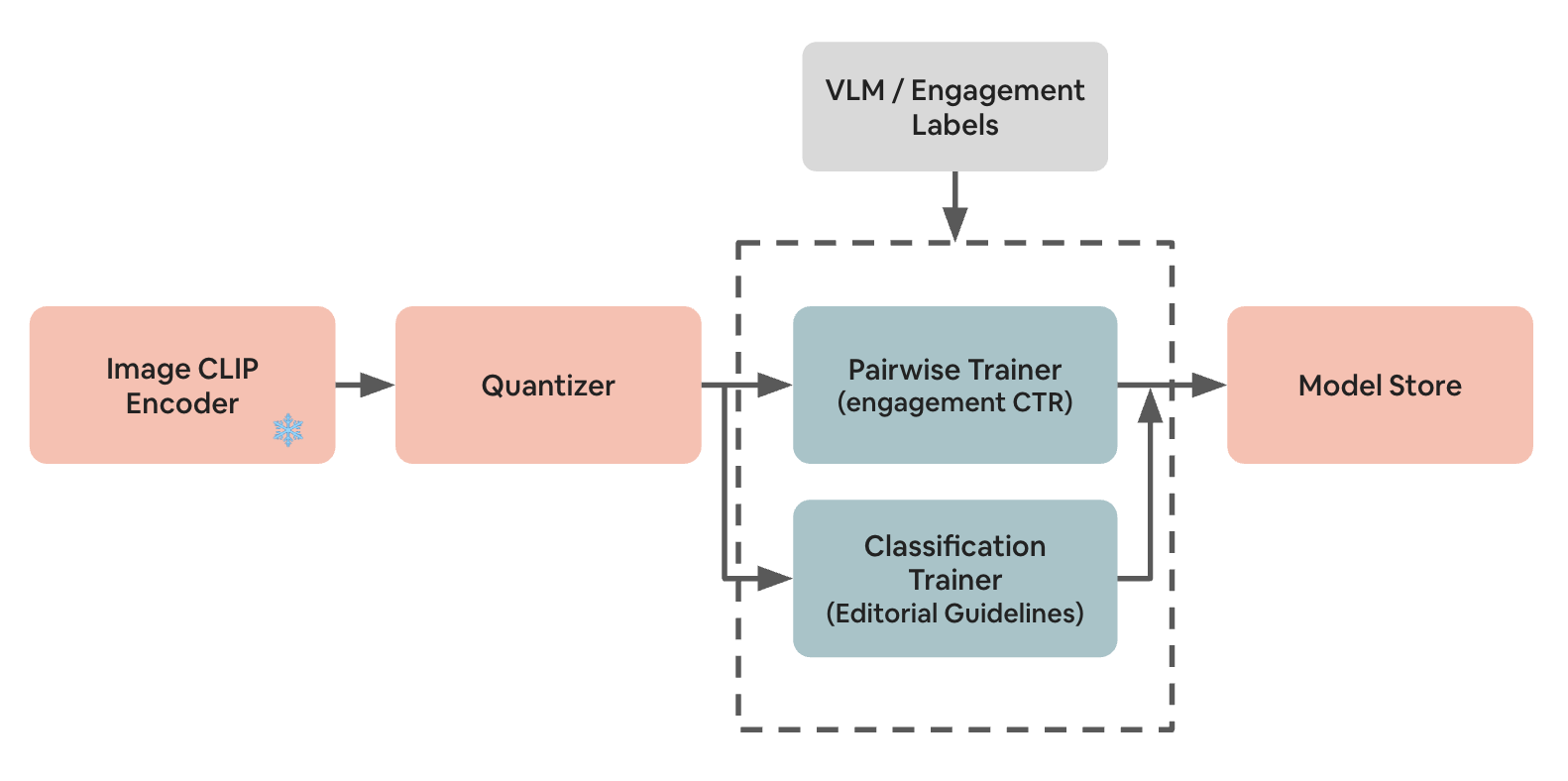}
    \caption{Vision heads framework}
    \Description[Vision heads framework Diagram]{A diagram showing the Vision heads framework with CLIP embeddings feeding into the quantizer, then undergoing either model training (classification or pairwise classification) before uploading the weights into the model store}
    \label{fig:vision-heads-Framework}
    \vspace{-20pt}
\end{figure}

\subsubsection{Data Preparation}
All tasks share the same vision backbone and associated quantizer for ease of use. The labels are collected using either a commercial VLM such as OpenAI's ChatGPT or from the online exploration (see Section~\ref{sec:Online-Policy-Engine}) in the case of the engagement model. 

\subsubsection{Training Details}

All vision heads are trained using the AdamW optimizer ($lr=0.001$ over 20 epochs) and a cross entropy loss. We use an early stopping criterion using a validation set. Then, all evaluations were performed on a test set. All training datasets were shuffled each epoch.

The CTR prediction vision head adopts a Siamese architecture trained with a pairwise learning-to-rank objective (Algorithm~\ref{algo:ctr-training}). Pairs of image embeddings are processed by a shared network and optimized to predict which image exhibits higher CTR. Importantly, this pairwise loss yields scores calibrated as the probability that an image wins, rather than as absolute CTR estimates.

\subsubsection{CTR Calibration}

Calibrated estimates of CTR for images is a desirable property of the CTR model.

Instead of calibrating the CTR logits during pairwise training which may hurt ranking performance, we trained a post-hoc calibration model. As the CTR model's primary aim remains to rank images, care must be taken to ensure that post-hoc calibration methods do not substantially alter the ordering. We explored isotonic and spline calibration methods as both methods guarantee monotonic behavior. In addition, spline calibration offers smooth transition thanks to its piecewise polynomial function. This property is particularly important in our setting, where the predicted CTR distribution is compressed and many images have similar predicted CTR values. Under these conditions, isotonic calibration tends to collapse neighboring scores into flat segments, eliminating fine-grained distinctions between images and reducing effective score resolution. In practice, while our experiments showed that isotonic and spline calibration achieved nearly identical calibration error, we selected spline calibration.

\subsubsection{Offline Results}

Table~\ref{tab:head-results} provides an overview of the performance across tasks. The performance contrasts positively with CLIP's well known zero-shot capabilities, especially in multilabel cases where zeroshot predictions require a specific arbitrary cutoff.

\begin{table}[H]

\centering
\begin{tabular}{l l c c}
\toprule
\textbf{Task} & \textbf{Metric} & \textbf{Baseline} & \textbf{Head (Ours)} \\
\midrule
Room type* & macro F1  & 0.54 & \textbf{0.96} \\ 
Annotation & macro F1 & 0.53 & \textbf{0.84}  \\ 
Human in photo & macro F1 & 0.66 & \textbf{0.92} \\ 
Amenities & macro F1 & 0.65 & \textbf{0.90} \\ 
HomePage CTR & Pairwise acc. & 57\% (Host) & \textbf{68\%} \\
\bottomrule
\end{tabular}
\caption{Vision heads performance, baseline is zero-shot unless specified otherwise, * indicates a multi-label task}
\vspace{-10pt}
\label{tab:head-results}
\end{table}

\subsection{Candidate Generation Engine}
\label{sec:candidate-generation-engine}
The objective of the Photo Candidate Selection is to generate cover photo candidates. To that end, we rely on proxy signals, such as whether a host has already used the photo as the cover in the past, or whether a photo is semantically similar to the one already vetted, to enqueue candidates. The human vetted pool was generated as part of our past work on Airbnb Categories HomePage, where we trained Vision Transformers to categorize photos into one of the Airbnb Categories, such as Lakefront, Countryside, Amazing Pools, Amazing View, etc. The candidates that were highly scored by those models were inspected by humans, who judged the photo quality and accepted or rejected the photo as cover photo for the specific category. Details can be found in the blog post \cite{Grbovic23}. This resulted in a large number of labels we could re-use in this project.

The candidate generation engine generated several waves of candidates that were sequentially vetted through online experiments. Table~\ref{tab:photo-candidates} provides a list of sources used to scale the pool of candidates. 

\begin{table}[H]
\centering
\begin{tabular}{l p{0.55\columnwidth}}
\toprule
\textbf{Source} & \textbf{Description} \\
\midrule
Current Host Cover & The currently selected host cover \\
Labeler Selected Cover & A photo selected by an internal labeler as part of the Airbnb Categories launch \cite{airbnb22} \\
Past Host Cover & A cover photo used in the past \\
Similar & A photo semantically similar to an already vetted candidate \\
Opportunistic Cover & A photo with high engagement improvement ($\Delta\hat{\mathrm{CTR}} \times \mathrm{weekly\_traffic}$) \\
\bottomrule
\end{tabular}
\caption{Sources of photo candidates}
\vspace{-20pt}
\label{tab:photo-candidates}
\end{table}
\begin{algorithm}[t]
\small
\KwIn{Frozen quantizer $Q$; pairwise data $D_{pair}$; calibration data $D_{cal}$; vision head $h$}
\KwOut{Trained head $h$ and calibrators $g_{s,c}$}

\textbf{Pairwise training.}
Freeze $Q$ and optimize only $h$ with AdamW.
\ForEach{batch $(codes_l,codes_r,w,s,c,CTR_l,CTR_r)\in D_{pair}$}{
    $z_l \gets \texttt{Gather}(h(Q.decode(codes_l)),s,c)$\;
    $z_r \gets \texttt{Gather}(h(Q.decode(codes_r)),s,c)$\;
    $p_{l\succ r}\gets\sigma(z_l-z_r)$\;
    $\ell \gets w\cdot\texttt{BCE}(\mathbbm{1}[CTR_l>CTR_r],p_{l\succ r})$\;
    Update $h$ using $\nabla_{\theta_h}\ell$\;
}

\textbf{Post-hoc calibration.}
Freeze $h$ and $Q$.
\ForEach{surface-context pair $(s,c)$ in $D_{cal}$}{
    $z_i \gets \texttt{Gather}(h(Q.decode(codes_i)),s,c)$ for all $i:s_i=s,c_i=c$\;
    $g_{s,c}\gets\texttt{FitMonotoneSpline}(\{(\sigma(z_i),CTR_i)\})$\;
}

\Return $h,\{g_{s,c}\}$\;
\caption{Pairwise CTR vision-head training with surface-context calibration.}
\label{algo:ctr-training}
\vspace{-12pt}
\end{algorithm}

\subsection{Listing Grounding Validator}
\label{sec:listing-grounding-validator}
    
The Listing Grounding Validator is a cross-modal validation module designed to assess whether a candidate photo is semantically representative of the underlying listing. Given the importance of visual misrepresentation in trip quality, the validator operates as a conservative filtering stage that prioritizes recall over precision to maintain trip quality.

For each candidate photo, the validator compares visual content against multiple sources of listing context, including host-provided captions and the full listing description (Table~\ref{tab:grounding-inputs}). The task is formulated as a binary grounding decision: whether the photo is consistent with the listing’s described space, amenities, and nearby points of interest, or whether it depicts content that is potentially misleading (e.g., unrelated landmarks, nearby attractions, or off-premise location).

The validation rubric is given below:
\begin{itemize}
    \item Is the photo of this property?
    \item Is the photo consistent with the listing facts?
    \item Does the photo truthfully represent the amenities, furnishings, and conditions the guest will experience?
\label{sec:listing-grounding-validator-rubric}
\end{itemize}

\begin{table}[tb]
\centering
\begin{tabular}{l p{0.65\columnwidth}}
\toprule
\textbf{Data} & \textbf{Details} \\
\midrule
Photo & The image content of the listing photo. \\
Caption & Host-provided caption, which may include contextual or descriptive voice-over (e.g., “This tavern is a 5-minute walk away”). \\
Listing description & Host-provided description of the space, including features, amenities, and nearby points of interest. \\
\bottomrule
\end{tabular}
\caption{Listing grounding validator input}
\vspace{-25pt}
\label{tab:grounding-inputs}
\end{table}

We implement the validator using a large multimodal language model (VLM) prompted to perform structured cross-modal classification over the image and textual inputs. The model produces a grounding verdict along with interpretable rationales, which are subsequently consumed as hard filters by the downstream ranking engine~\ref{sec:photo-ranking-engine}.

To evaluate, a team of experts annotated 1k photo–listing pairs. Based on the performance benchmark, we selected OpenAI’s 4o-mini model for its strong recall characteristics, achieving a macro F1 score of 0.82 within this benchmark. This recall-oriented operating point aligns with our product requirement to minimize unrepresentative candidates, ensuring that potentially misleading photos are filtered before being surfaced to guests.

\subsection{Candidate Ranking Engine}
\label{sec:photo-ranking-engine}

The Candidate Ranking Engine converts the candidate pool (Section~\ref{sec:candidate-generation-engine}) into a context-aware ranking for online selection. Given a listing $l$ and context $(s,c)$ (e.g., surface, query, or filters), the engine applies:

\begin{enumerate}[noitemsep, topsep=0pt]
    \item \textbf{Quality filtering:} remove candidates failing visual or editorial policies using the vision heads (Section~\ref{sec:vision-heads}).
    
    \item \textbf{Representativeness validation:} apply the Listing Grounding Validator (Section~\ref{sec:listing-grounding-validator}) to higher-risk candidates to reject misleading or off-premise images.
    
    \item \textbf{Eligibility and ranking:} assign contextual tags (e.g., \texttt{pool}, \texttt{lakefront}) and retain only candidates matching the surface intent. Eligible candidates are ranked by calibrated predicted CTR.
\end{enumerate}

Formally, the eligible set is
$\mathcal{X}=\{x:q(x)=1,\;r(x,l)=1,\;E(x,c,s)=1\}$,
where $q$ denotes quality filtering, $r$ representativeness validation, and $E$ contextual eligibility. Candidates are scored by
$\text{score}(x \mid c,s)=\hat{CTR}(x,c,s)$
and sorted in descending order.

The resulting ranked list defines the action space used by the Exploration Logging Module (Section~\ref{sec:Online-Policy-Engine}).

\subsection{Exploration Logging Module (ELM)}
\label{sec:Online-Policy-Engine}
OptiCover operates in a significant cold-start regime where most candidate photos have no prior engagement labels. The primary role of the online component is therefore twofold: (i) collect unbiased engagement feedback over a dynamically constructed candidate set, and (ii) deploy a stable selection policy that adapts to heterogeneous product surfaces and context.

We evaluated a hybrid Thompson Sampling policy as an online baseline, but found that in our ultra-cold-start and large-action-space setting, the calibrated exploit policy grounded in offline engagement modeling initially performed substantially better. The Thompson Sampling policy gradually closed this gap as observations accumulated (Section~\ref{subsubsection:Online-policy-comparison}), suggesting that its initial underperformance was driven by slow convergence under sparse and noisy rewards, further compounded by system feedback lag. This motivated our use of the calibrated offline model as the production exploit policy, particularly for newly introduced candidates with limited engagement history.

\subsubsection{Exploration and Logging}
For each listing and context, the Candidate Ranking Engine (Section~\ref{sec:photo-ranking-engine}) produces a daily set of $k$ eligible photo candidates, along with eligible context, which defines the action space. To collect unbiased training data, an online $\epsilon$-greedy exploration policy with $\epsilon=0.1$: with probability $\epsilon$ used to uniformly sample an eligible candidate; otherwise, we serve the current exploit choice. For each impression, we log the candidate set, the selected action, surface and context features, and downstream engagement events. This produces randomized, balanced, comparisons across candidate photos, which we later convert into photo pairs for the engagement head.

\subsubsection{Deployment Policy (Exploit)}
In exploitation, we select the eligible candidate with the highest calibrated predicted CTR (Section~\ref{sec:vision-heads}) for the context $c$ using the output of the photo ranking engine \ref{sec:photo-ranking-engine}, stored daily in a Key Value store. Since the CTR model is lightweight, retraining and deployments carry a low operational burden.

\subsection{Online Evaluation}
\label{sec:online-eval}
To confirm top-of-the-funnel (HomePage) engagement improvement as well as booking improvement on the search page, we test OptiCover in 3 A/B tests and report the online results in Table~\ref{tab:online-results}. Across all experiments, the control group is the current host covers in all A/B tests. 


\begin{table}[tb]
\centering
\small
\begin{tabular}{lcccc}
\toprule
\textbf{Surface} &
\textbf{Duration} &
\textbf{Aprx. N} &
\textbf{CTR lift} &
\textbf{Bookers lift} \\
\midrule

HomePage
& 3 wks
& $\sim$10M
& \makecell{\textbf{+1.5\%}\\{\scriptsize [1.3\%, 1.8\%]}}
& -- \\

\makecell[l]{Search\\(Contextual Exploit)}
& 6 wks
& $\sim$1M
& \makecell{+0.14\%\\{\scriptsize [-0.24\%, 0.52\%]}}
& \makecell{\textbf{+0.35\%}\\{\scriptsize [0.01\%, 0.69\%]}} \\

\makecell[l]{Search\\(non-contextual)}
& 2 wks
& $\sim$10M
& \makecell{\textbf{+0.32\%}\\{\scriptsize [0.16\%, 0.48\%]}}
& \makecell{+0.075\%\\{\scriptsize [-0.14\%, 0.29\%]}} \\

\bottomrule
\end{tabular}

\caption{Online experiment results. $N$ denotes the approximate number of
unique visitors. Values report relative treatment effects with 95\%
confidence intervals. Bold indicates statistical significance at
$\alpha=0.05$.}
\label{tab:online-results}
\end{table}

All experiments use visitor-level randomization between the current host-selected cover (control) and OptiCover (treatment). Treatment effects are reported as relative changes from control with 95\% confidence intervals, computed using CUPED when available.

For HomePage and Search, we evaluate an Exploit policy that selects candidates using predicted CTR. The CTR model is trained using pairwise engagement data collected by the ELM~\ref{sec:Online-Policy-Engine} over a 5-week exploration period. Due to data availability, the deployed model is trained exclusively on HomePage data rather than using the generalized multi-surface training procedure described in Algorithm~\ref{algo:ctr-training}. This provides a natural test of cross-surface transfer.

On Search, we first evaluated a non-contextual variant that directly applies this transferred CTR model to select the highest-scoring eligible photo. Despite producing a statistically significant 0.32\% CTR lift, this treatment produced no significant increase in Bookers (+0.075\%, 95\% CI: $[-0.14\%, 0.29\%]$). Analysis indicated that the additional engagement primarily redistributed bookings across listings rather than generating incremental bookings. This result highlights an important limitation of directly optimizing an intermediate engagement metric on a downstream marketplace surface: increased local engagement does not necessarily translate into incremental conversion.

In parallel, we evaluated a contextual Search policy that constrains the candidate action space according to the guest's search intent before applying the engagement model. Opticover is only applied to seven filters selected by the Product team: pools, hot tub, kitchen, waterfront, lakefront, ski-in/ski-out, and indoor fireplace. In contrast to the non-contextual policy, the contextual treatment produces a significant 0.35\% increase in Bookers (95\% CI: $[0.01\%, 0.69\%]$), despite a non-significant +0.14\% CTR effect. Together, these experiments demonstrate that action-space contextualization can be more important than maximizing the intermediate CTR objective itself when optimizing downstream marketplace outcomes.

The Search experiments also demonstrate a practical form of cross-surface transfer: a CTR model trained entirely on HomePage engagement data can be used to bootstrap optimization on a new surface before sufficient surface-specific training data becomes available. Algorithm~\ref{algo:ctr-training} generalizes this approach to surface-specific engagement models once such data is available.

Finally, we report Bookers because it is Airbnb's primary metric for revenue attribution. The observed Search improvement corresponds to millions of dollars in incremental revenue.
\subsection{Ablation Study}
\label{sec:ablation-study}

To understand the role played by the different components of this complex workflow, we conduct ablation studies on the different components:
\begin{itemize}
    \item \textbf{Ablation of Candidate Generation Engine} focusing on how candidate generation, through its critical expansion of the action space affects performance
    \item \textbf{In-training Auxiliary calibration loss vs. post training calibration} justifying the added complexity by superior calibration results
    \item \textbf{Ablation of the Listing Grounding Validation} showing higher representativeness than unimodal approaches
    \item \textbf{Online Policy Comparison} showing that a strong offline CTR exploit policy outperforms Thompson Sampling during cold start, while the decreasing regret gap suggests that Thompson Sampling improves as online observations accumulate

\end{itemize}

\subsubsection{Ablation of Candidate Generation Engine}
\label{subsubsection:Ablation-of-Candidate-Generation-Engine}

In this ablation, we focus on the relationship between the action space formed by adding candidate sets of cover photos and the CTR on HomePage simulated using an action-space oracle analysis. We first remove all sources of candidates (Table~\ref{tab:photo-candidates}) with only host-selected photos remaining in the action space, then add one source at a time, and compute the CTR of each listing by taking the CTR of the best performing photo from the action space for that listing.

We can observe the improvement of CTR as additional sources are added. Metrics are reported in Table~\ref{tab:Action-space-vs-model-performance}.

\begin{table}[tb]
\centering
\begin{tabular}{lcc}
\toprule
\makecell{\textbf{Candidate photo set}\\\textbf{added (cum.)}} 
& \makecell{\textbf{Best CTR}\\\textbf{(normalized)}} 
& \makecell{\textbf{Photo count}\\\textbf{(normalized)}} \\
\midrule
Current Host Covers & 1.00 & 1.00  \\
+ Labeler Selected Covers & 1.10 & 1.19 \\
+ Past Host Covers & 1.23 & 1.34  \\
+ Similar & 1.34 & 1.47 \\
+ Opportunistic Covers & 1.38 & 1.51 \\
\bottomrule
\end{tabular}
\caption{CTR of best performing photo improves with additional candidates}
\label{tab:Action-space-vs-model-performance}
\end{table}

In conclusion, having strong candidates in the action space is a primary driver of performance in semantically constrained environment.

\subsubsection{In-training auxiliary calibration vs.\ post-hoc calibration}
\label{subsubsection:ctr-regression-vs-pairwise}

We evaluated two approaches for CTR calibration.

\begin{figure}[tb]
  \centering
  \includegraphics[width=\linewidth]{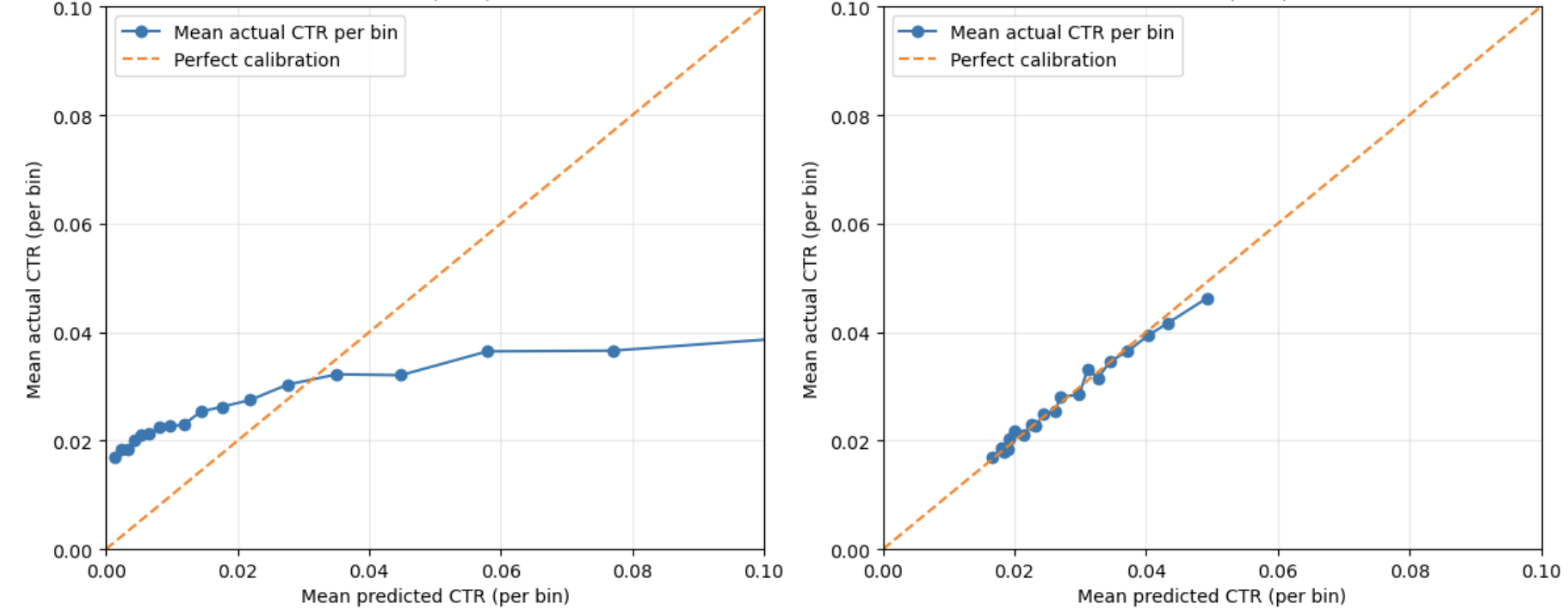}
  \caption{Comparison of in-training auxiliary calibration loss (left) and post-training calibration (right) on a held-out set.}
  \Description{Calibration comparison}
  \label{figure:calibration-comparison}
\end{figure}

First, following Wang et al.~\cite{wang21} Eq.~11, we augmented the pairwise ranking loss with a pointwise MSE calibration term weighted by $\gamma$. Hyperparameter sweeps showed improved raw CTR calibration, with $\gamma=0.42$ providing the best tradeoff between calibration and ranking quality. However, calibration errors remained, particularly at the extremes (Figure~\ref{figure:calibration-comparison}).

Second, we applied a post-hoc monotonic calibrator (spline regression). Although this adds an offline calibration step, it consistently reduced miscalibration more than the auxiliary in-training loss alone, motivating our final approach.

\subsubsection{Ablation of the Listing Grounding Validator}
\label{subsubsection:ablation-of-the-listing-grounding-validation}

We evaluated the impact of the Listing Grounding Validator on 1k randomly sampled candidates produced by the Candidate Engine. Candidates were annotated for representativeness using the rubric in Section~\ref{sec:listing-grounding-validator-rubric}.

We compare the full pipeline against a ``no-grounding'' variant relying only on engagement ranking, filtering, and a vision-only representativeness classifier. Removing the grounding validator reduced recall of non-representative photos from 98\% to 89\%, resulting in a \textbf{5.3$\times$ increase} in non-representative outputs while filtering 14\% fewer candidates. Most failures were off-premise images (e.g., nearby POIs).

These results show that grounding validation is critical for preventing misleading photo selection.
\subsubsection{Online Policy Comparison}
\label{subsubsection:Online-policy-comparison}

We compare two online selection policies on HomePage: Thompson Sampling (TS) and the CTR Exploit policy. In an 18-day online experiment, TS is initialized with 8 weeks of prior online experiment traffic and then updated daily at the photo level, with its Beta posterior parameters $\alpha$ and $\beta$ updated using impressions and clicks observed on the previous day.

We report the regret gap between TS and CTR Exploit. Because both policies are evaluated against the same oracle for a given candidate set, the oracle term cancels, and the regret gap is equivalent to the difference between the CTRs of the photos selected by CTR Exploit and TS:
\begin{equation}
\Delta R
=
R_{\mathrm{TS}} - R_{\mathrm{Exploit}}
=
\operatorname{CTR}(x_{\mathrm{Exploit}})
-
\operatorname{CTR}(x_{\mathrm{TS}}).
\end{equation}
Positive values therefore indicate that CTR Exploit selects a higher-CTR photo than TS.

\begin{figure}[tb]
\centering
\includegraphics[width=\linewidth]{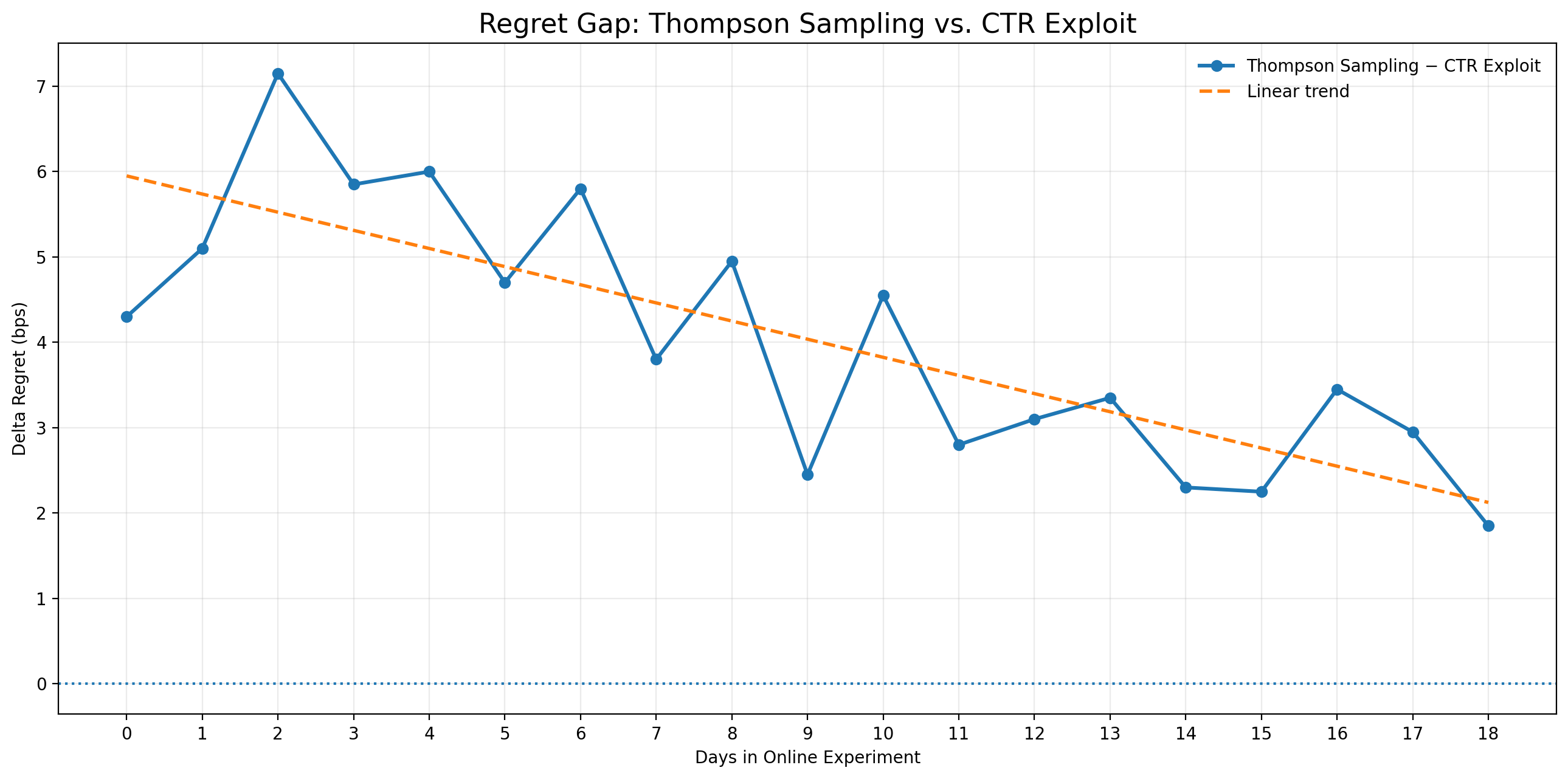}
\caption{Regret gap between Thompson Sampling and CTR Exploit over
the online experiment. Positive values indicate better performance
for CTR Exploit.}
\label{fig:ts-vs-ctr-exploit}
\end{figure}

CTR Exploit maintains a positive regret gap throughout the experiment. However, the gap decreases by approximately a factor of two over the 18-day period, indicating that TS improves as online observations accumulate. Despite this narrowing gap, TS does not match the performance of the offline-trained CTR Exploit policy within the experimental window. This result highlights the practical value of a strong offline prior when online rewards are sparse and noisy.

\section{Future Work}

Moving forward, the work can be extended from contextualization to personalization by incorporating guest-level preferences, as different guest segments may resonate with different photo types (e.g., families preferring pool photos in summer), building on contextual bandit methods~\cite{Li10}. 

Another important direction is addressing a current system limitation: the inability to change the order of listing cards. Studying the benefits of integrating photo merchandising into the listing-card ranking model—jointly ranking listings and photos—may further enhance business impact.

\section{Conclusion}

We proposed OptiCover, an approach to optimize contextual optimal cover photos in the industrial context of Airbnb listings. We demonstrated the usefulness of the approach across product surfaces, and objective functions. The framework introduces new components: a Listing Ground Validator for candidate validation, and a vision head framework for scalable vision tasks. The framework provides a useful compass to companies exploring photo optimization to drive conversion and ultimately revenue gains. 

Looking forward, OptiCover sets the foundations for future work as the personalization can be extended to more guest need-states and, in the end, to the individual guest themselves.

\section*{GenAI Usage Disclosure}
OpenAI’s ChatGPT was used to assist with brainstorming, red-teaming, grammar checking, and rephrasing during the preparation of this manuscript. 

The authors retain full responsibility for the accuracy, originality, and integrity of all content in this work.

\section*{Acknowledgements}
The authors thank the various contributors to this project for direct and indirect support. In particular, Sherry Chen, Soumyadip Banerjee, Pu Yang, Lu Zhang, and Zhenyu Lai, Phanindra (Phani) Ganti as well as the executive support of Sanjeev Katariya, Dean Chen, and Kedar Bellare.


\clearpage
\bibliographystyle{ACM-Reference-Format}
\bibliography{citations}
\balance


\end{document}